\documentclass[twoside,leqno,twocolumn]{article}

\usepackage[letterpaper]{geometry}
\usepackage{amsmath,amssymb}
\usepackage{algorithm, algorithmic}
\usepackage{graphicx, subfigure}
\usepackage{booktabs}
\usepackage{threeparttable}
\usepackage{booktabs}
\usepackage{ltexpprt}

\begin{document}

\title{\Large Contextual Local Explanation for Black Box Classifiers}
\author{Zijian Zhang \and Fan Yang \and Haofan Wang \and Xia Hu}

\date{}

\maketitle


\fancyfoot[R]{\scriptsize{Copyright \textcopyright\ 2020 by SIAM\\
Unauthorized reproduction of this article is prohibited}}





\begin{abstract} \small\baselineskip=9pt 
We introduce a new model-agnostic explanation technique which explains the prediction of any classifier called CLE. CLE gives an faithful and interpretable explanation to the prediction, by approximating the model locally using an interpretable model. We demonstrate the flexibility of CLE by explaining different models for text, tabular and image classification, and the fidelity of it by doing simulated user experiments.
\end{abstract}

\section{Introduction}

Complex machine learning models are widely utilized in a myriad of applications such as text classification\cite{TextClassification}, image recognition\cite{ImageClassification} and so on, even though the complexity of these models makes them black box which are hard for people to understand. However, it is important for people to understand what the model is doing and why it predicts so, especially when people deploy it in business or industry. As a result, interpretable machine learning\cite{IML} has become hot spot these years, people have proposed many approaches to interpret behavior of the model. In this paper, we focus on local methods based on perturbation, such as LIME\cite{LIME}. and Anchors\cite{Anchors}.

LIME can explain the predictions of any classifier or regressor in a faithful way, by approximating it locally with an interpretable model. LIME explanation is easy to interpret and friendly to understand. However, the coverage of it is unclear. In other words, the LIME explanation can be somewhat unstable. For example, sometimes a weight for one word can be positive in one instance but negative in another. The explanation is unclear, which makes people confused and perplexed. For example, LIME explanations for two sentiment predictions made by an LSTM\cite{LSTM} in Figure 1(b). Although both explanations are computed to be locally accurate, if one takes the explanation on the left and tries to apply it to the sentence on the right, one might be tempted to think that the word ``not" would have positive influence, but it does not.

Anchors, another model-agnostic system based on if-then rules. An anchor explanation is a rule that sufficiently ``anchors" the prediction locally, which is always composed of combination of words. For instances which satisfy the rule, there are chances that the prediction is always the same. Anchor explanation's coverage is very clear and precise with much less human effort involved, because people will not get involved in computing the contribution of the features in linear explanations. Nevertheless, Anchor explanation can be limited sometimes. For example, since its coverage is too narrow, it only applies to instances that satisfy the rule. For another thing, although Anchor ensure the local prediction by using combination of words, it does not give numeric data to express how exactly they contribute to the prediction like LIME does. For example, explanations for two sentiment predictions in Figure 1(c). They are very straight-forward and understandable, but for users who are sensitive to exact numbers, they are not that reliable.

\begin{figure}[t]
\centering

\subfigure[Instances]{
\begin{minipage}[b]{0.5\textwidth}
\includegraphics[scale=0.36]{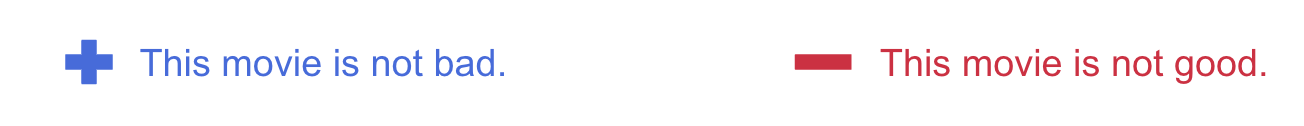}
\end{minipage}
}

\subfigure[LIME]{
\begin{minipage}[b]{0.5\textwidth}
\includegraphics[scale=0.23]{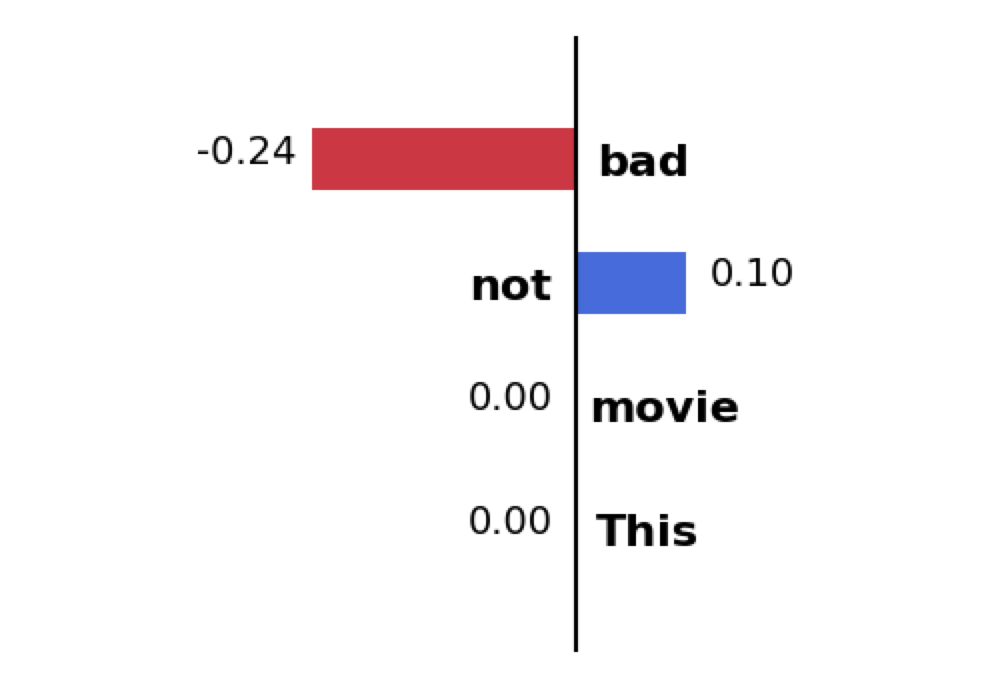}
\includegraphics[scale=0.23]{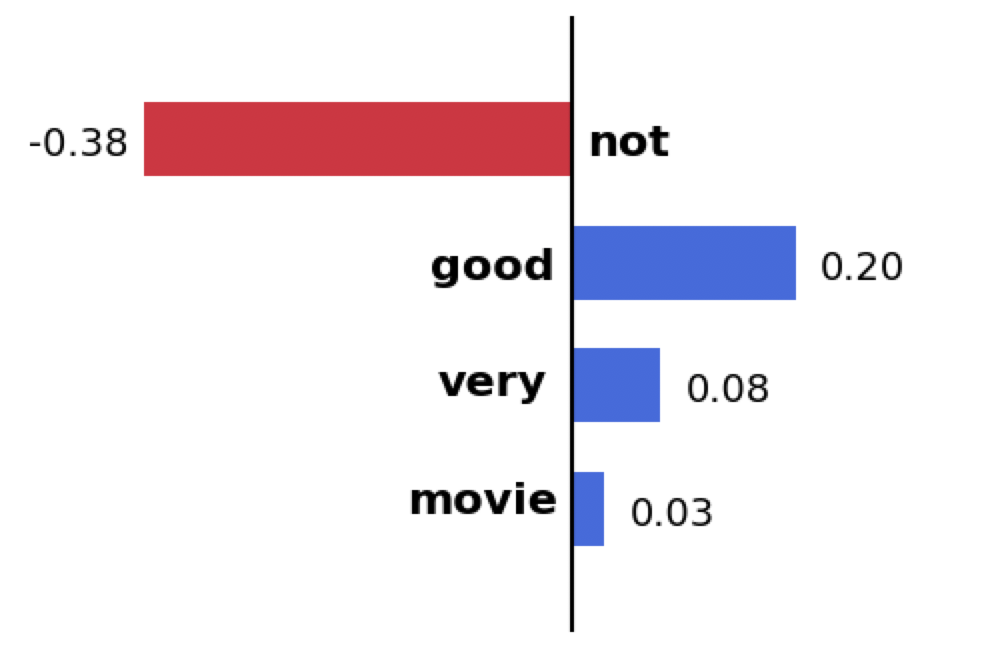}
\end{minipage}
}

\subfigure[Anchors]{
\begin{minipage}[b]{0.5\textwidth}
\includegraphics[scale=0.31]{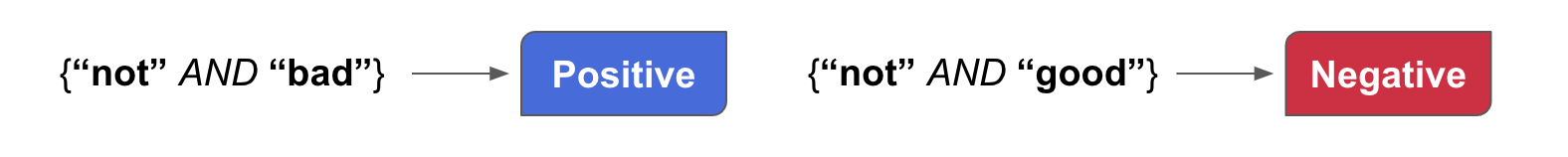}
\end{minipage}
}

\subfigure[CLE]{
\begin{minipage}[b]{0.5\textwidth}
\includegraphics[scale=0.21]{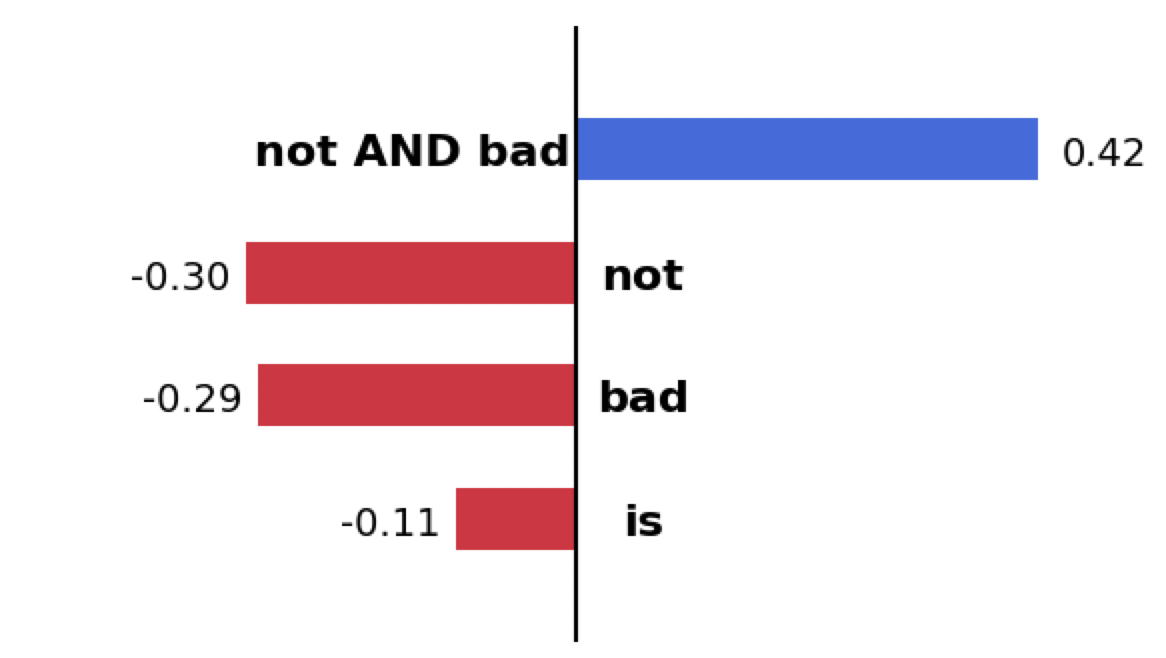}
\includegraphics[scale=0.21]{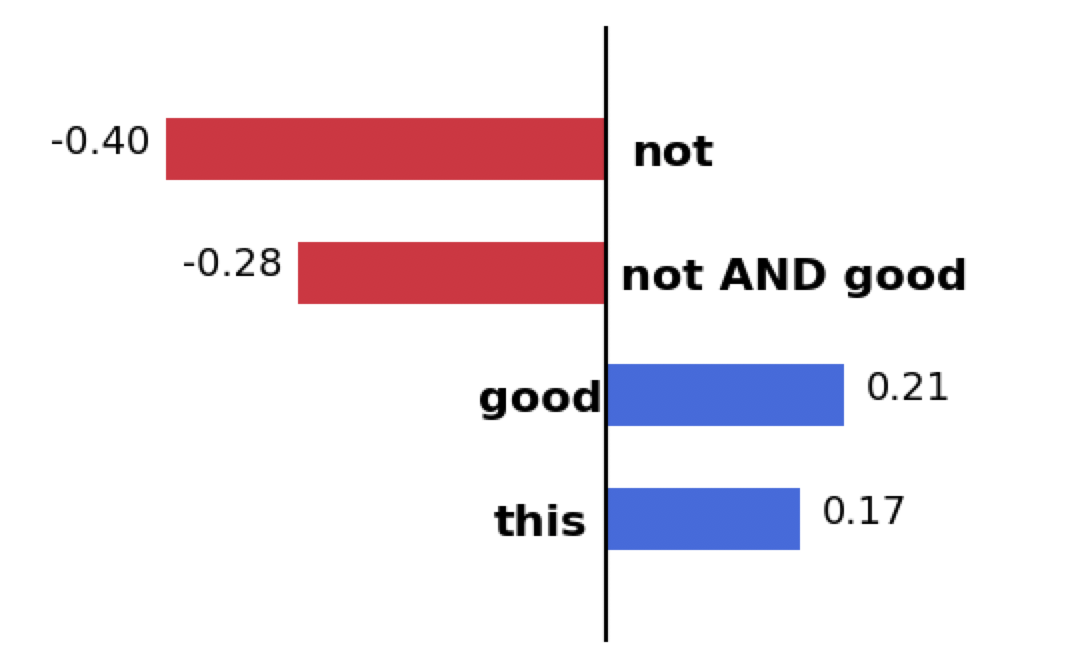}
\end{minipage}
}

\caption{Explanations}
\end{figure}

\begin{figure*}[t]
    \centering
    \includegraphics[scale=0.5]{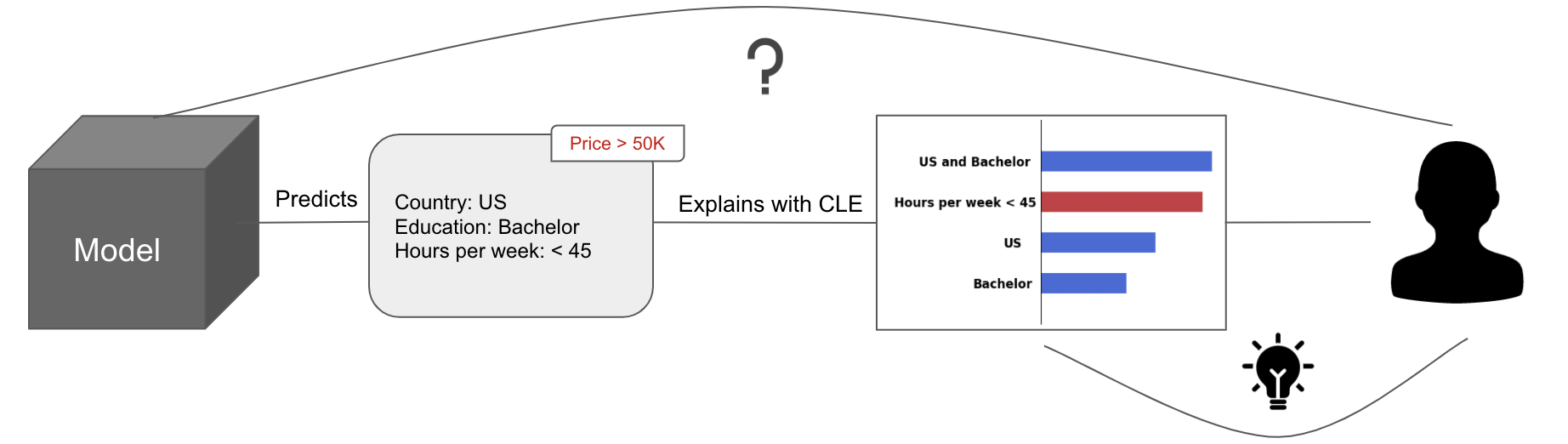}
    \caption{Overall View}
\end{figure*}

How can we combine the best of both of these worlds? In this paper, we introduce a novel model-agnostic explanation called CLE. CLE can explain the prediction in a more interpretable way by both taking combination of features into account and giving numeric data as result. CLE explanation contains relationship between feature and feature. As a result, the explanation is stable, straight-forward and interpretable. For instance, explanations for two sentiment predictions in Figure 1(d). The weights of word ``not" and ``bad" separately are both negative, but the weight of ``not AND bad" is positive. Then, people can have some intuitive thoughts, ``not" and ``bad" separately have negative effect to the result, but when ``not” and ``bad" appear in the mean time, they can have positive effect. 

CLE helps people spot a light into the black box model, then people can have some insights like how the model predicts, and whether the model is reliable. The overall view is shown in Figure 2. We demonstrate the usefulness of CLE by applying it to diversified domains(text, tabular and images), and the fidelity of it by doing user simulated experiments.

\section{Preliminary}

Before we dive into the detail of CLE, we will introduce several basic conceptions.

\subsection{Interpretable data representations}

Different machine learning models have different types of data as their input. For example, people may input raw text, image or tabular data into the model. Consequently, in model-agnostic explanation task, people need to find a uniform interpretable data representation which is understandable to human. For example, a possible interpretable representation in text task can be a N-dimension binary vector indicating the presence or absence of a feature, regardless of what feature the black box model use. Likewise, we can also apply this vector to image task. Since it is computationally difficult to denote a whole bunch of pixels, we use the vector to store the presence of each super pixel in the image, while the classifier may use tensor to denote it. To conclude, when the black box classifier use $x \in  R^d$ as input, then we use $x^{'} \in \{0, 1\}^d$ to denote it.

\subsection{Perturbation}

In model-agnostic explanation task, the inner part of the model is invisible. When explains an instance, we always perturb\cite{Perturbation} the instance locally, and use these created samples to test the behavior the model. Assume that the instance is denoted by a vector $x^{'} \in \{0, 1\}^d$, we perturb around $x^{'}$ by setting ones in $x^{'}$ to zeros uniformly at random. Given a perturbed instance $p^{'} \in \{0, 1\}^d$, we can recover the black box representation $p \in R^d$, with which we can obtain the model output \textit{R} by inputting it into the black box model.

\subsection{Fidelity-Interpretability Trade-off}

The most common way to achieve interpretability is to use interpretable models, such as linear regression, decision tree, to approximate the black box model locally. We define an explanation as model $m \in M$ where \textit{M} is the whole family of interpretable models which accept $p^{'} \in \{0, 1\}^d$ as input. And let \textit{f} be the black box model which takes $p \in R^d$ as input and outputs \textit{R}. We need to ensure the fidelity of model \textit{m}, which means the local model outputs $R^{'}$ are supposed to be close to \textit{R}. Thus, we use $\pi_{x}$ as a measure of the proximity between instance $\textit x^{'}$ and perturbed data $p^{'}$, and then fit model \textit{m} using data ${p}^{'}$ and black box model output \textit{R} with weight $\pi_{x}$.

The interpretability of models $m \in M$ is different. We use $\Omega(m)$ to denote the difficulty of a model to interpret. For example, in decision tree model, $\Omega(m)$ can be the number of branches or the depth of the tree. Normally, high fidelity means relatively high complexity, thus low interpretability, and vice versa. Hence, we need to trade off between them and choose the best model \textit{m}. Let ${L}(f, m, {\pi_{x}})$ be the measure of how unfaithful that \textit{m} is the local approximation of \textit{f} with $\pi_{x}$. In order to achieve the balance between fidelity and interpretability, we are supposed to minimize \textit{L} and keep $\Omega(m)$ low enough for human to understand.

\begin{equation}
\theta(x) = \operatorname*{argmin}_{m \in M} \textit{L}(f, m, {\pi_{x}}) + \Omega(m)
\end{equation}

This formulation can be applied to any model in \textit{M}. In this paper, we focus on linear explanation.

\subsection{LIME}

Let \textit{M} be the class of linear models, such that $m(p^{'})=w_{m}\cdot p^{'}$. LIME use the weighted squared loss as \textit{L}, and let $\pi_{x}(p) = exp(-D(x,p)^{2}/\sigma^{2})$ be an exponential kernel function defined on D with width $\sigma$.

\begin{equation}
{L}(f, m, {\pi_{x}}) = \sum_{p,p^{'} \in P} \pi_{x}(p)(f(z)-m(p^{'}))
\end{equation}

For text classification, they ensure the interpretability by defining the interpretable representation of bag of words. And they set a limit K on the number of words to control the complexity $\Omega(m)$. Likewise, they use the same $\Omega$ for image classification, they use ``super-pixel" instead of words, such representation of an image is a binary vector where 1 indicates the original pixel, while 0 denotes grayed out or average super-pixel. And in tabular data classification, 1 stands for the original data, 0 means to discretize the column by method such as quartile.
After learning the weight(using Algorithm 2.1) via least square procedure and Lasso\cite{Lasso}, we can get K weights. LIME explains the instance by assigning K features with their corresponding weights. Thus, we can have an intuition on how these features affect the prediction.

\begin{algorithm}
\caption{LIME}
\begin{algorithmic}
\REQUIRE Classifier \textit{f}, number of samples \textit{N}
\REQUIRE Instance \textit{x}, interpretable version $x^{'}$
\REQUIRE Similarity kernel $\pi_x$, length of explanation K
\STATE $P \gets \{\}$
\FOR{$i \in \{1,2,3...N\}$}
\STATE $p_{i}^{'} \gets perturb(x^{'})$
\STATE $P \gets P \cup (p_{i}^{'}, f(p_{i}), \pi_{x}(p_{i}))$
\ENDFOR
\STATE $\omega \gets K-Lasso(P, K)$ with $p^{'}$ as features, $f(p)$ as targets
\RETURN $\omega$
\end{algorithmic}
\end{algorithm}

\section{CLE}

Although LIME explanation is easy to interpret, the coverage of it is unclear, resulting unstable weight on the same feature. Chances that people will be confused when a feature's weight is positive in one instance but negative in another. Notice that interpretable representation in LIME only takes the presence of individual feature into account, one weight corresponds to one feature. As a result, the explanation is incapable to capture the relationship regarding the combination of several features, and that's why the explanation can be unstable. We now introduce the process of CLE concretely. CLE proposes a new form of interpretable representation which takes combination of features into account and changes the process of perturbation. Finally, it improves the drawback in LIME and gets a more interpretable explanation.

\subsection{Interpretable representations with combination}

In LIME, original data $x \in  R^d$ is represented by $x^{'} \in \{0, 1\}^d$, while 0 stands for absence and 1 denotes presence of the word. Now suppose that we want to take the presence of the combination of every 2 words into account, we can extend $x^{'} \in \{0, 1\}^d$ by $x^{e} \in \{0, 1\}^{C_{d}^2}$, and the extended part of this binary vector is used to represent the simultaneous presence of certain pair of features in $x^{'}$. For example, $x^{e}(1)$ indicates the simultaneous presence of $x^{'}(1)$ and $x^{'}(2)$, $x^{e}(2)$ indicates the presence of $x^{'}(1)$ and $x^{'}(3)$ \dots Hence, our data representation is capable to include the information with regard to combination of features.

\begin{figure}[h]
    \centering
    \includegraphics[scale=0.4]{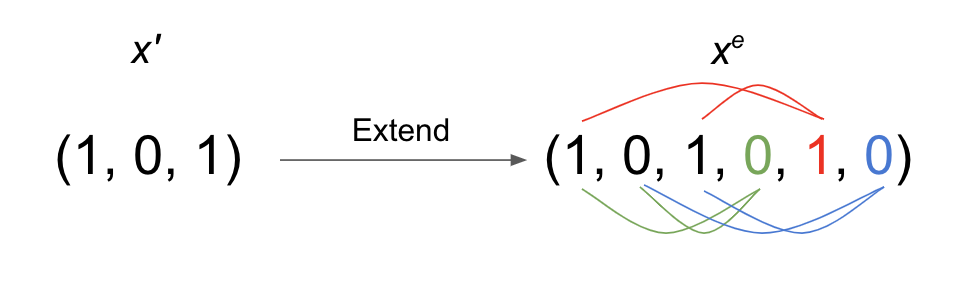}
    \caption{Interpretable Representations with Combination}
\end{figure}

\subsection{Perturbation and extension}

Suppose the instance we want to explain is denoted by a vector $x^{'} \in \{0,1\}^d$, we then perturb around $x^{'}$ locally by drawing zero nonzero elements of $x^{'}$ uniformly random. Given a perturbed instance $p^{'} \in \{0,1\}^d$, we can calculate the extended extended binary vector $p^{e} \in \{0,1\}^l$. Actually, the combination of features is not only limited to pair, we can also take the combination of three or more features into account, let the spans of combination users care about be a vector $B^{\mu}$, for example, (2,3..). However, too many combinations will include some meaningless ones that users do not care about, so we define a vector $E^{\lambda}$ which contains the index of features that users care about. The length of extended binary vector $p^{e}$ is determined by $B^{\mu}$ and $E^{\lambda}$. After applying extension algorithm(Algorithm 3.1) to perturbed data $p^{'}$, the dimension of it changes from \textit{d} to $l = \sum_{i=1}^{\mu} C_{B^{\mu}(i)}^{\lambda}$.

\begin{algorithm}
\caption{Extension with combinations}
\begin{algorithmic}
\REQUIRE perturbed data $p^{'}$
\REQUIRE a list of combinations $B^{\mu}$, a list of indexes $E^{\lambda}$
\STATE $N \gets \sum_{i=1}^{\mu} C_{B^{\mu}(i)}^{\lambda}$
\STATE $p^{e} \gets ones(N)$
\STATE $combs = [\ ]$
\FOR{$b \in \{B^{\mu}(1),B^{\mu}(2)...B^{\mu}(\mu)\}$}
    \STATE $extend(combs, Combinations(E, b))$
\ENDFOR
\FOR{$n \in \{1,2,3,...N\}$}
    \STATE $comb \gets combs(n)$
        \FOR{$item \in comb$}
            \IF{$p^{'}(item)\ \textbf{is}\ 0$}
                \STATE $p^{e} \gets 0$
                \STATE \textbf{break}
            \ENDIF
        \ENDFOR
\ENDFOR
\RETURN $extend(p^{'}, p^{e})$
\end{algorithmic}
\end{algorithm}

\subsection{Sparse linear explanation}

In CLE, we have $m(p^{e})=w_{m}\cdot p^{e}$. And we also use the weighted squared loss as \textit{L}, and let $\pi_{x}(p) = exp(-D(x,p)^{2}/\sigma^{2})$ be an exponential kernel function defined on D with width $\sigma$.

\begin{equation}
{L}(f, m, {\pi_{x}}) = \sum_{p,p^{e} \in P} \pi_{x}(p)(f(z)-m(p^{e}))
\end{equation}

\begin{algorithm}[h]
\caption{{CLE}}
\begin{algorithmic}
\REQUIRE Classifier \textit{f}, number of samples \textit{N}
\REQUIRE Instance \textit{x}, interpretable version $x^{'}$
\REQUIRE Similarity kernel $\pi_x$, length of explanation K
\STATE $P \gets \{\}$
\FOR{$i \in \{1,2,3...N\}$}
\STATE $p_{i}^{'} \gets perturb(x^{'})$
\STATE $P \gets P \cup (p_{i}^{'}, f(p_{i}), \pi_{x}(p_{i}))$
\STATE $p_{i}^{e} \gets extend\_with\_combination(p_{i}^{'})$
\ENDFOR
\STATE $\omega \gets K-Lasso(P, K)$ with $p^{e}$ as features, $f(p)$ as targets
\RETURN $\omega$
\end{algorithmic}
\end{algorithm}

For example, in text classification, we ensure the interpretability by defining the extended interpretable representation of bag of words. We always limit $B^{\mu}$ to (2,\ ), in that the combination of two words always matters much, and we set indexes $E^{\lambda}$ to the words we take interests in, and number of features we care about to K. It can not only control the complexity $\Omega(m)$, but also makes explanation targeted. When explains an instance, we first perturb locally around the instance, and extend the perturbed data with Algorithm3.1. Then we make use of the extended perturbed interpretable data to fit a linear regressor. After that, we explains the instance by assigning weights to corresponding features as their importance. Figure 4 gives an intuition about what the CLE is doing.

\begin{figure}[t]
    \centering
    \includegraphics[scale=0.25]{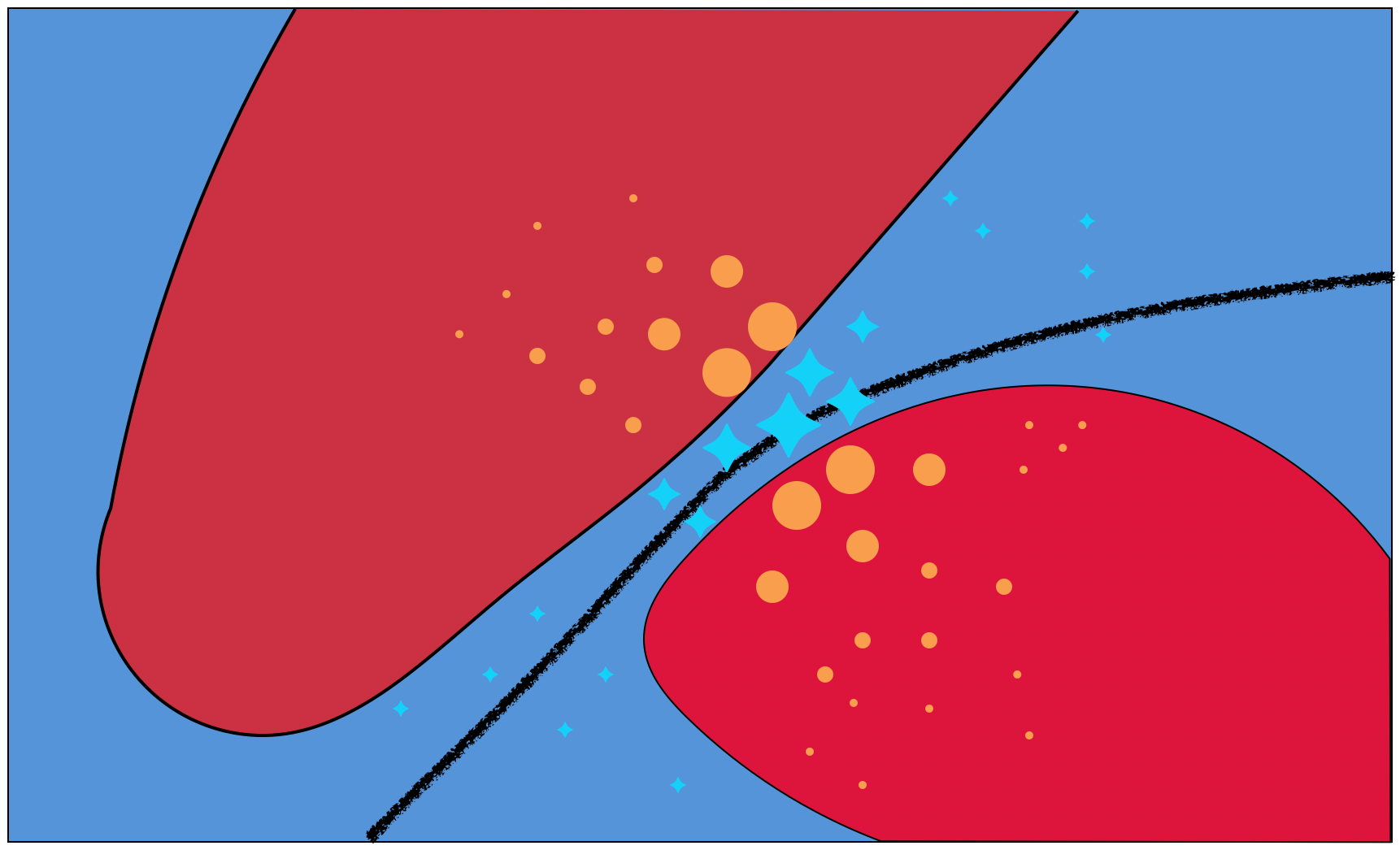}
    \caption{toy example}
\end{figure}

\subsection{Analysis of complexity}

Assume that we want to explain an instance $x \in  R^d$ with \textit{N} perturbed samples, we focus on features with indexes $E^\lambda$, and we only care about combinations of every \textit{2} features. First, we use $x^{'} \in \{0, 1\}^d$ to represent \textit{x}, perturb $x^{'}$ locally \textit{N} times, and then we can get perturbed data $p^{'}$. We define $t(d)_{pert}$ to denote the time it takes to perturb, thus this process will take $N*t(d)_{pert}$ time. Second, we pass perturbed data \textit{p} corresponding to $p^{'}$ into black box model \textit{f} to get \textit{N} labels. If it takes the model $t(p)_{pred}$ time to predict once, then this step will spend $N*t(p)_{pred}$ time. After that we need to extend $p^{'}$ with combinations to get $p^e$. Since we only care about combinations of \textit{2} features, it will take $N*t(\lambda^{2})_{ext}$ time. Next, we need to fit the local linear model with $p^e$ as features and $f(p)$ as targets, we define the time this procedure takes is $t(p^{e})_{fit}$ time. Finally, we can conclude the complexity of CLE.

\begin{equation}
\textit{T}(x) = N* [t(d)_{pert} + t(p)_{pred} + t(\lambda^{2})_{ext}] + t(p^{e})_{fit}
\end{equation}

Thus, when apply CLE to the black box model, we need to choose a suitable number of samples \textit{N} to balance between time complexity and local accuracy. And as we do experiments, we notice that when the black box model is fairly difficult, which means $t(p)_{pred}$, \textit{T} is dominated by the prediction process. Moreover, when the number of features of your dataset is big (nearly ten thousand), the process of extension $t(\lambda^{2})_{ext}$ and fit $t(p^{e})_{fit}$ can be very computational hard. Users can accelerate this process by choosing the feature they care about, which decreases the $\lambda$.

\subsection{Example 1: Text classification with RF}

In figure 6, we explain the prediction of a random forest classifier trained on \textit{sentence polarity dataset} to distinguish between ``negative" and ``positive" movie comments. This classifier only achieves 70\% accuracy, which means this classifier is not that reliable. People may probably pay much attention on the word ``blunt" or ``indictment" in this sentence, however, the explanation shows that the classifier pays much attention on the word ``film", ``of" and ``bring". 
Having get some intuitions from the explanation, it is obvious that this classifier is not trust-worthy. And maybe we should try another model that can better understand the structure of the sentence in the next step.

\begin{figure}[h]
    \centering
    \includegraphics[scale=0.47]{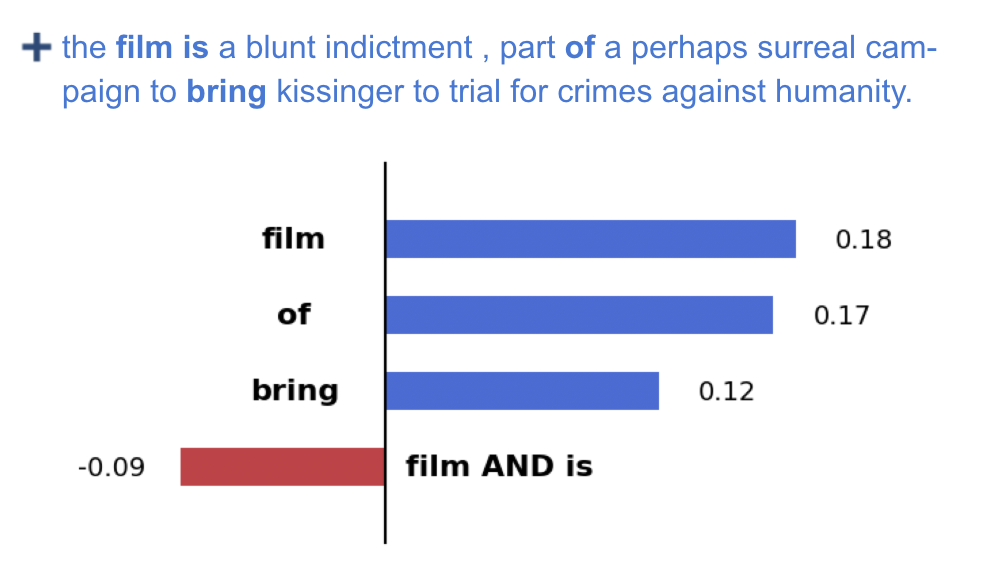}
    \caption{CLE Explanation for RF}
\end{figure}

\subsection{Example 2: Image classification with deep networks}

We also apply both CLE and LIME to a much more complex model, inception-v3\cite{Inception-V3}, in image classification task on dataset \textit{Imagenet}\cite{ImageNet}. We fill the positive-weight segments with green color, and red color vice versa. Moreover, the absolute value of weights are denoted by the depth of the color. The explanation of a random image is showed in Figure 7. Figure 7 shows the LIME explanation for the top 2 labels violin and cello. And it also shows the CLE explanation while setting $B^\mu$ to (2,). In this example, LIME explanation's coverage is not that precise, and the difference of weight between segments is not clear, while CLE clearly points out the exact part with differentiated weights. Having get the explanation, it is apparent that inception-v3 is a reasonable and reliable model.

\begin{figure}[h]
    \centering
    \includegraphics[scale=0.4]{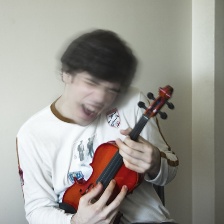}
    \caption{original image}
\end{figure}

\begin{figure}[h]
\centering
\subfigure[LIME]{
    \subfigure[violin]{
        \includegraphics[scale=0.3]{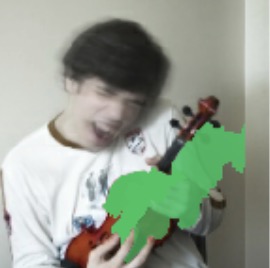}
    }
    \subfigure[cello]{
        \includegraphics[scale=0.3]{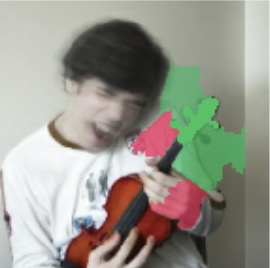}
    }
}
\quad
\subfigure[CLE]{
    \subfigure[violin]{
        \includegraphics[scale=0.3]{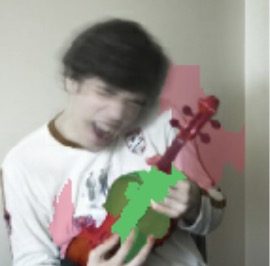}
    }
    \subfigure[cello]{
        \includegraphics[scale=0.3]{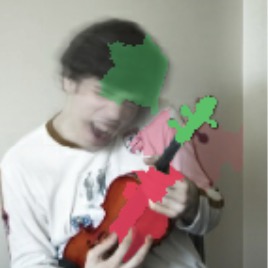}
    }
}

\caption{LIME and CLE Explanations}
\end{figure}

\begin{figure}[!htbp]
    \centering
    \includegraphics[scale=0.375]{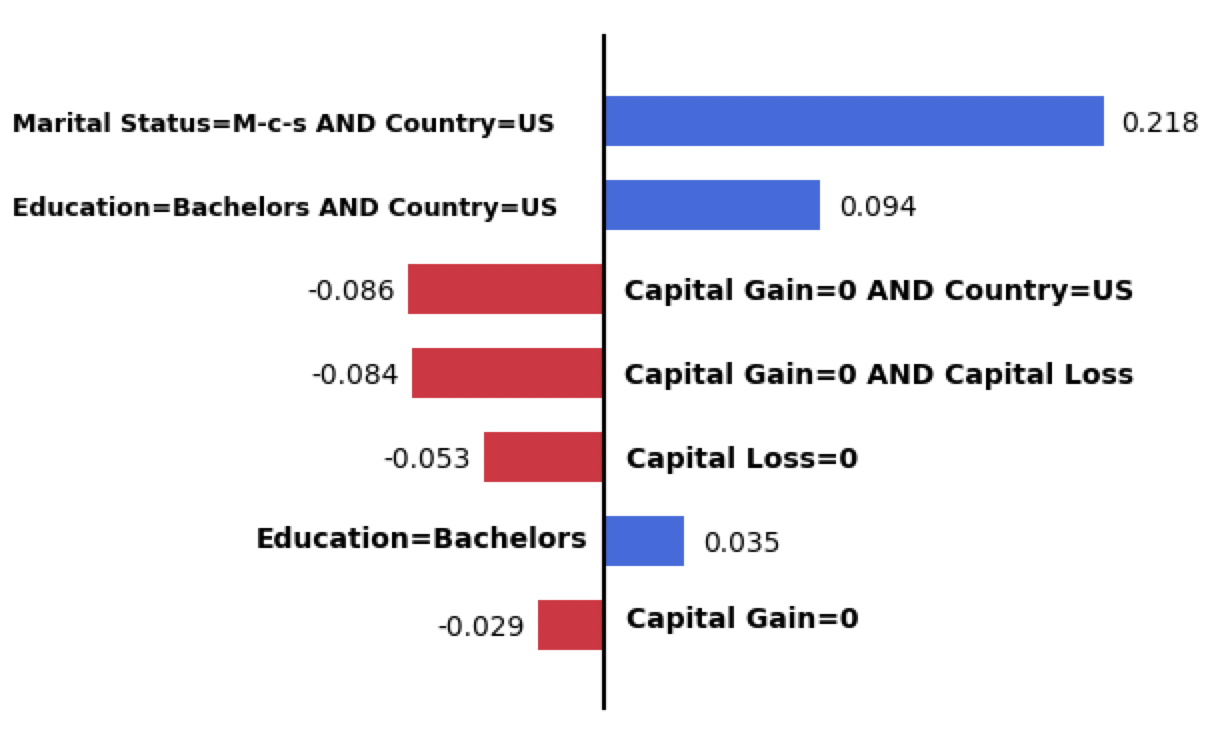}
    \caption{Tabular CLE Explanation}
\end{figure}

\subsection{Example 3: Tabular classification with SVMs}

As for tabular classification, we explain the prediction of a support vector machine with RBF kernel trained on \textit{adult dataset}. It predicts whether an adult can earn more than 50K per month according to his information, such as gender, hours per week, country and so on. This classifier achieves 85\% accuracy, and we explain a random row from test dataset, the result is shown in Figure 8. People can easily get some intuitions from these two explanations. In LIME explanation, Capital Gain, Capital Loss and Martial Status have much influence on the prediction. While in CLE explanation, Martial Status alone have little influence but it matters much when combined with Country=US, and Capital Loss or Capital Gain have much extra effect when combined with other features.

\section{Simulated User Experiments}
In this part, we will focus on the comparison between LIME and CLE. As what LIME does, we also use two sentiment analysis dataset (books and DVDs), where the target is to distinguish between negative and positive reviews\cite{Polarity}. We train a variety of models including decision tree(\textbf{DT}), logistic regression with L2 regularization(\textbf{LR}), nearest neighbor(\textbf{NN}) and support vector machine(\textbf{SVM}) with RBF kernel. We use the same parameters as LIME's experiments, and divide the data into train data(1600 instances) and test data(400 instances). Moreover, in order to further explore the difference between LIME and CLE, we also use ImageNet dataset, where the task is to classify the object in the image. We use the pretrained model Inception-V3, which is much more complex than the previous machine learning models. Since Imagenet dataset is so large, so we do the experiment with images randomly chosen from validation set. For LIME and CLE, we set the number of sample \textit{N} to 15000 for both methods, and set $B^\mu$ = (2,) to accelerate the computational process. And we set the number of maximum features \textit{K} to 10 in our experiments. Moreover, we compare CLE with a method called greedy\cite{Greedy} procedure, in which we iteratively remove the feature that contributes the most to the black box prediction for \textit{K} times. We also use a procedure called random, in which we randomly pick the top \textit{K} important features, to be the baseline.

Experiments can be divided into three parts. In the first part, we will test the faithfulness of the explanation to the black box model. Then, we are going to simulate a user and test whether the local model prediction is reliable. Finally, we simulate a scenario where users hesitate between two models with similar accuracy, and test if our explanation can help user to choose between them.

\subsection{Are explanations faithful to the model?}

We measure the faithfulness of the explanation with the models which are interpretable themselves, including logistic regression and decision tree. We train these two models with maximally 10 features, as a result, we can get the \textit{gold} set of features the model largely depends on. For each explanation on the test dataset, we compute the fraction of these \textit{gold} features that are covered by the explanations. We calculate the recall averaged over test dataset and the result is shown in Table 1. As a result, random procedure only gets pretty low recall. Greedy procedure obtains about 60\% recall on linear model but relatively lower on tree model. Both CLE and LIME achieve \textgreater 90\% recall for both classifiers on both dataset, CLE does better in logistic regression model, while LIME does better in decision tree model.  For models which are not interpretable themselves and the methods that have local prediction, we measure the faithfulness by calculating the absolute error between local prediction and model prediction. We then report this error in Table 2, 3 and 4. Consequently, the absolute error in CLE explanation is less than LIME explanation on all classifiers and datasets. Particularly, in models which are fairly complex, such as SVM and Inception-V3, CLE local prediction is much more nearest to the black box model's prediction. It gives us an intuition that when black box model is relatively difficult and nonlinear, CLE can capture the nonlinearity and better approximate the black box model.
 
\begin{figure}[h]
    \centering
    \includegraphics[scale=0.2]{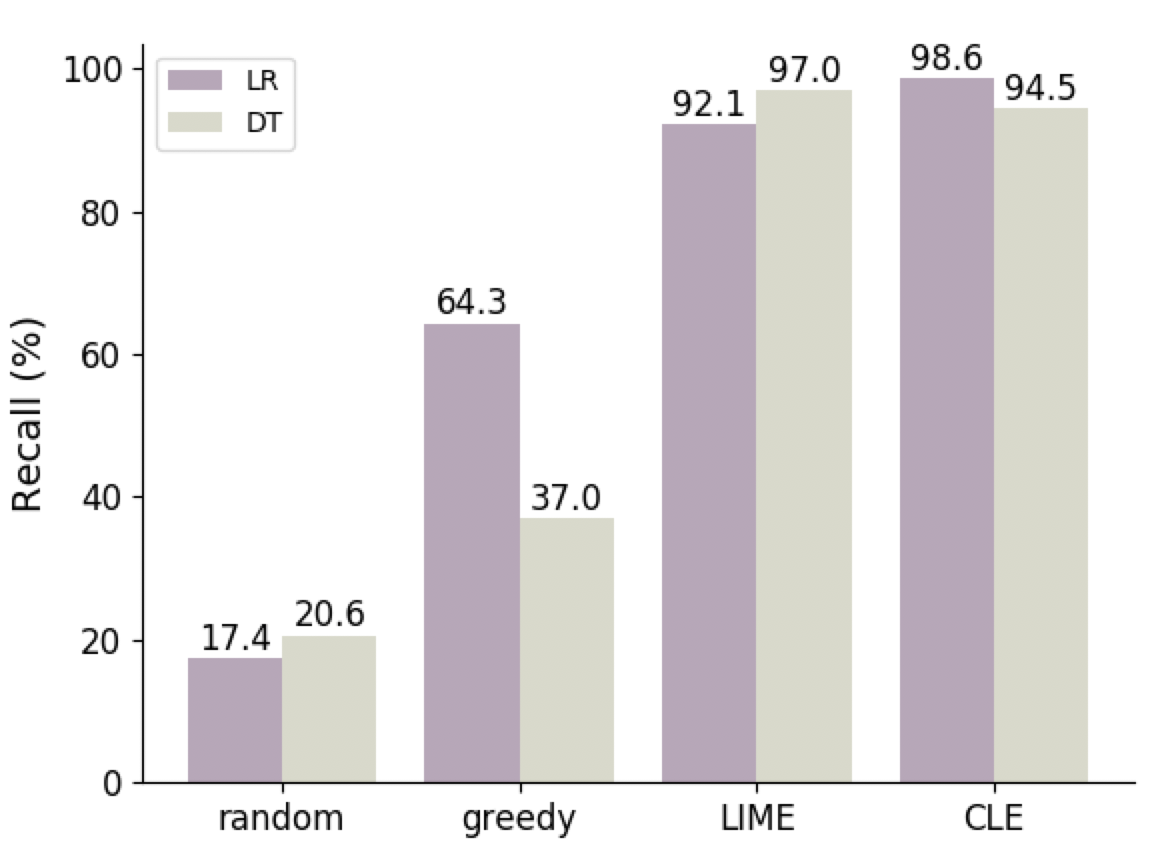}
    \caption{Recall on gold features on the books dataset}
\end{figure}

\begin{figure}[h]
    \centering
    \includegraphics[scale=0.2]{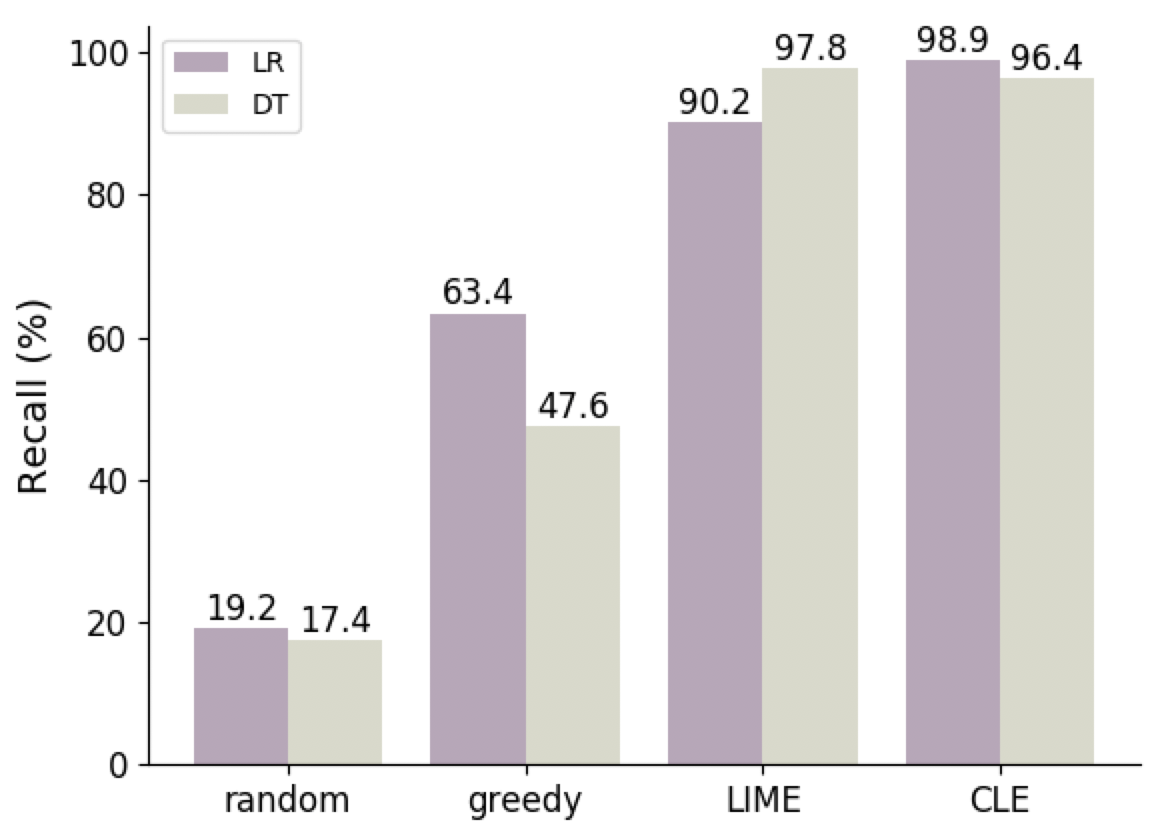}
    \caption{Recall on gold features on the DVDs dataset}
\end{figure}

\renewcommand{\arraystretch}{1.5}
\begin{table}[h]
 
  \centering
  \fontsize{10}{8}\selectfont
  \begin{threeparttable}
  \caption{Absolute error on Books dataset}
  \label{tab:performance_comparison}
    \begin{tabular}{ccccc}
    \toprule
    &LR&DT&NN&SVM\cr
    \midrule
    LIME& 0.0057 & 0.0932 & 0.0730 &  0.0257 \cr
    CLE & {\bf 0.0023} & {\bf0.0293} & {\bf 0.0544} & {\bf 0.0057}\cr
    \bottomrule
    \end{tabular}
    \end{threeparttable}
\end{table}

\renewcommand{\arraystretch}{1.5}
\begin{table}[h]
 
  \centering
  \fontsize{10}{8}\selectfont
  \begin{threeparttable}
  \caption{Absolute error on DVDs dataset}
  \label{tab:performance_comparison}
    \begin{tabular}{ccccc}
    \toprule
    &LR&DT&NN&SVM\cr
    \midrule
    LIME& 0.0087 & 0.0852 & 0.0772 &  0.0260 \cr
    CLE & {\bf 0.0027} & {\bf0.0266} & {\bf 0.0625} & {\bf 0.0065}\cr
    \bottomrule
    \end{tabular}
    \end{threeparttable}
\end{table}

\renewcommand{\arraystretch}{1.5}
\begin{table}[!htbp]
 
  \centering
  \fontsize{10}{8}\selectfont
  \begin{threeparttable}
  \caption{Absolute error on ImageNet dataset}
  \label{tab:performance_comparison}
    \begin{tabular}{ccc}
    \toprule
    &
    LIME & CLE\cr
    \midrule
    Inception-V3& 0.2564 & {\bf 0.0283}\cr
    \bottomrule
    \end{tabular}
    \end{threeparttable}
\end{table}

\subsection{Should I trust the prediction?}

Should users trust the prediction of the local model? In order to simulate trust in predictions, we first randomly select 25\% of the features (in image task, we randomly choose 10\% segments) to be ``unrelated", which means the simulated user will not trust these features. We then develop ``trustworthiness" by labeling test dataset predictions from black box model as ``untrustworthy" if the prediction changes when ``unrelated" features are removed from the instance, and label the prediction ``trustworthy" otherwise. We assume the simulated user will deem the LIME or CLE predictions ``untrustworthy" if the prediction from the approximated linear model changes after remove the ``unrelated" features. As a result, we can test whether the simulated user will trust the prediction.

We report the f1 score on the trustworthy predictions for each explanation in Table 5, 6 and 7. As a result, for book and DVDs dataset, the average f1 scores are approximately similar in LIME and CLE. However, again, for extremely complex model like deep networks, LIME explanation is not reliable as CLE explanation.

\renewcommand{\arraystretch}{1.5}
\begin{table}[!htbp]
 
  \centering
  \fontsize{10}{8}\selectfont
  \begin{threeparttable}
  \caption{Average F1 of trustworthiness on Books dataset}
  \label{tab:performance_comparison}
    \begin{tabular}{ccccc}
    \toprule
    &LR&DT&NN&SVM\cr
    \midrule
    Random&14.6&14.8&14.7&14.7 \cr
    Greedy&53.7&47.4&45.0&53.3 \cr
    LIME& 96.6 & 97.9 & {\bf94.5} & {\bf96.7} \cr
    CLE & {\bf96.7} & {\bf99.2} & 94.2 & 96.3\cr
    \bottomrule
    \end{tabular}
    \end{threeparttable}
\end{table}

\renewcommand{\arraystretch}{1.5}
\begin{table}[!htbp]
 
  \centering
  \fontsize{10}{8}\selectfont
  \begin{threeparttable}
  \caption{Average F1 of trustworthiness on DVDs dataset}
  \label{tab:performance_comparison}
    \begin{tabular}{ccccc}
    \toprule
    &LR&DT&NN&SVM\cr
    \midrule
    Random&14.2&14.3&14.5&14.4 \cr
    Greedy&52.4&58.1&46.6&55.1 \cr
    LIME& {\bf96.6} & 98.4 & {\bf91.8} & 95.6 \cr
    CLE & 95.2 & {\bf98.5} & 90.2 & {\bf95.7}\cr
    \bottomrule
    \end{tabular}
    \end{threeparttable}
\end{table}

\renewcommand{\arraystretch}{1.5}
\begin{table}[h]
 
  \centering
  \fontsize{10}{8}\selectfont
  \begin{threeparttable}
  \caption{Average F1 of trustworthiness on Imagenet dataset}
  \label{tab:performance_comparison}
    \begin{tabular}{ccccc}
    \toprule
    &
    Random&Greedy&LIME & CLE\cr
    \midrule
    Inception-V3& 57.8 & 58.2 & 90.6 & {\bf 92.7}\cr
    \bottomrule
    \end{tabular}
    \end{threeparttable}
\end{table}

\subsection{Can it help with choice between models?}

In this part, we will evaluate whether our explanation is helpful with model selection. Assume that a user gets two models with similar validation accuracy, and we will test whether our explanation can do a favor to him. To simulate this scenario, we add 10 artificially ``noisy" features. Concretely, for training and validation sets, each artificial feature appears in 10\% of the examples in one class, and 20\% of the other. While on the test instances, each artificial features appears in 10\% of the examples in each class. Consequently, the dataset becomes a mixture of informative features and ``garbage" features which introduce spurious correlations. We create a pair of random forests with 30 trees until their validation accuracy is within 0.1\% each other, but test accuracy differs by at least 5\%. In this situation, it is impossible to choose the better classifier according to the validation accuracy.
After getting \textit{P} instances from the validation set, and thus \textit{P} explanations, the simulated human marks the artificial features that appear in the \textit{P} explanations as untrustworthy, and then we can evaluate how many predictions for the validation set can be trusted. After that, we select the classifier with less untrustworthy explanations as the better one, and check if this classifier is the one with higher test set accuracy.
Since the accuracy of random procedure is always wander around 50\%, we only show the result of other three methods. The result is shown in Figure 11 and 12. The accuracy of greedy procedure is relatively lower.  CLE and LIME approximately achieve the same accuracy. When the number of instances is relatively smaller, CLE does a little better, when instances get bigger, LIME gives a better result.

\begin{figure}[h]
    \centering
    \includegraphics[scale=0.333]{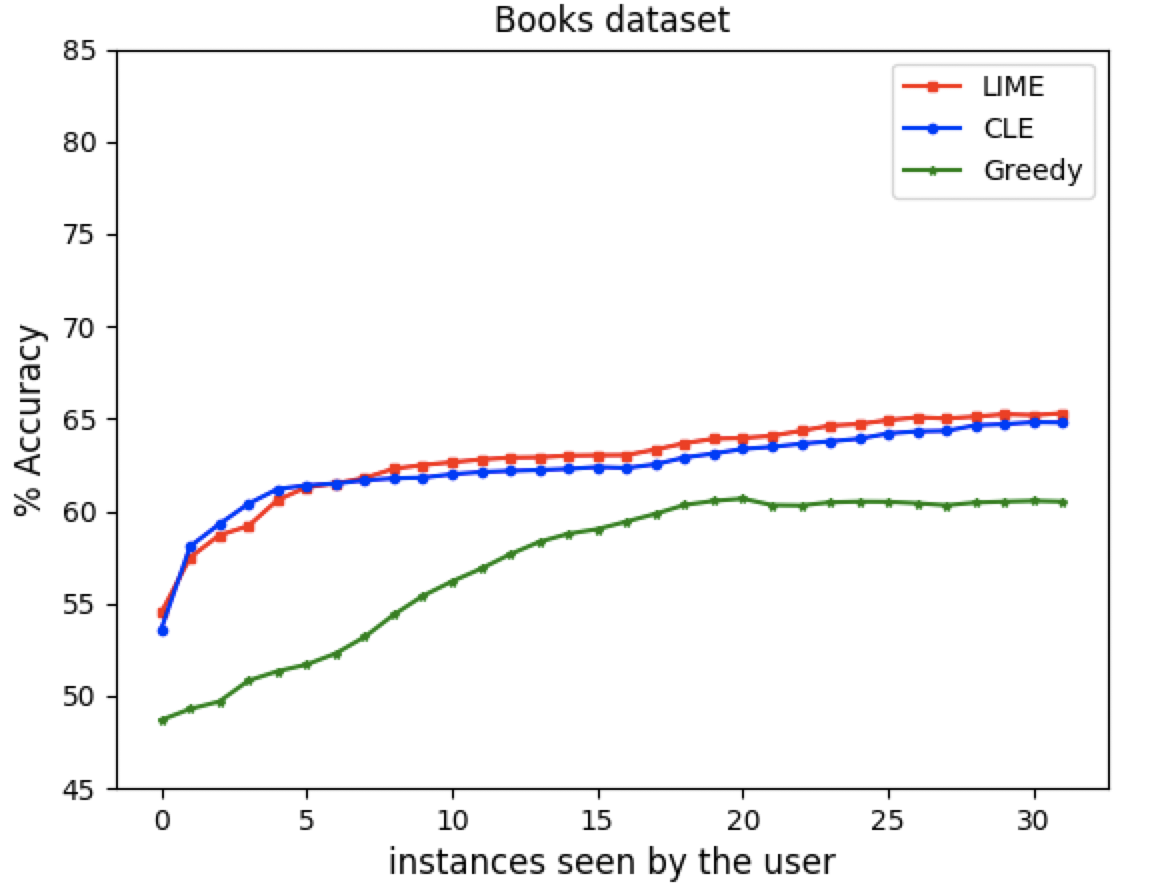}
    \caption{Accuracy choosing between two classifiers on Books dataset}
\end{figure}

\begin{figure}[h]
    \centering
    \includegraphics[scale=0.333]{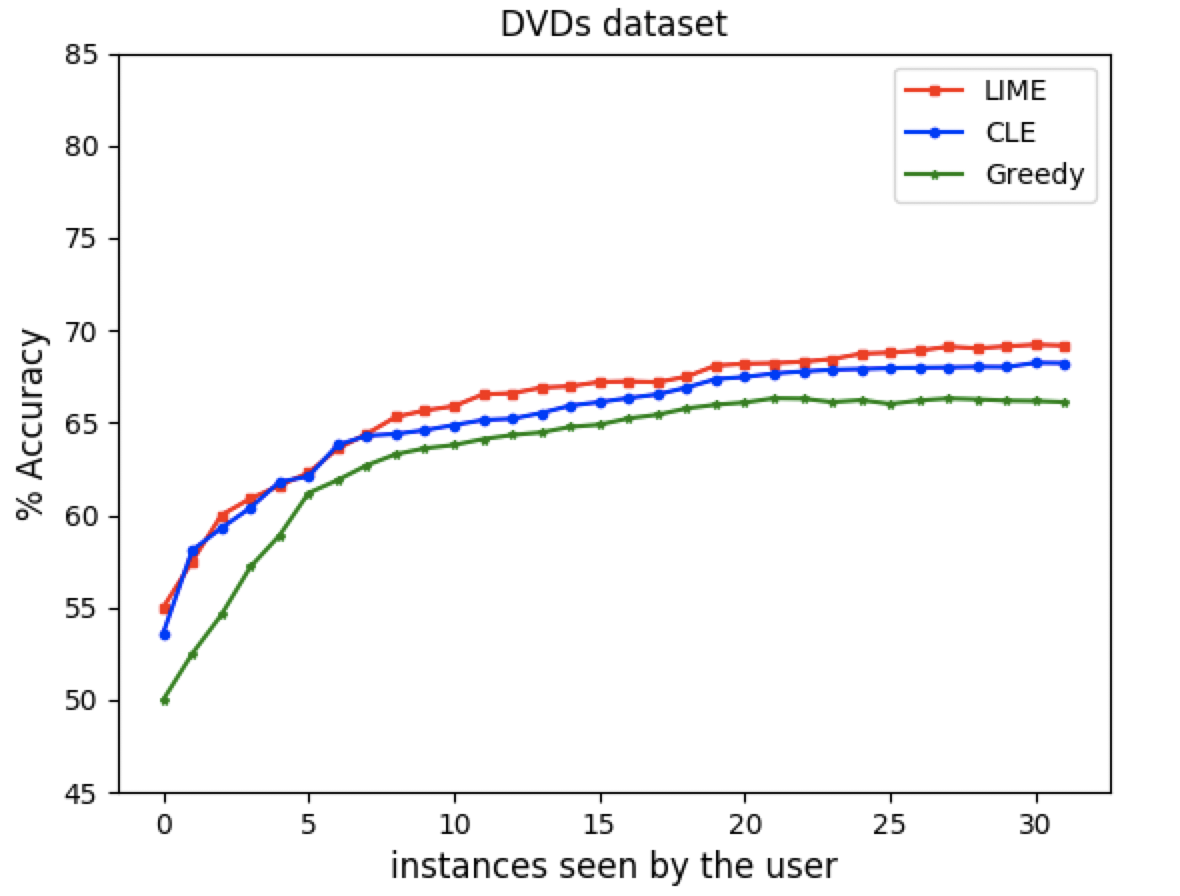}
    \caption{Accuracy choosing between two classifiers on DVDs dataset}
\end{figure}

\section{Related work}

We divide the methods which explain the model into two parts. For one thing, the methods based on perturbation. People perturb the instance to be explained locally, and use these perturbed samples to test the behavior of the model, finally get some intuitions. The methods based on perturbation, besides LIME and Anchors, also include SHAP\cite{SHAP}. Shapley Value is a game theory conception. SHAP (SHapley Additive exPlanations), which takes advantage of Shapley Value, is a unified approach to explain the output of any model. It provides the additive feature attribution based on expectations. Moreover, SHAP not only has the model-agnostic kernel explainer, but also includes model-specific explainers, such as tree explainer and deep explainer.

For another thing, there are also many methods based on gradients. These methods are usually used in image tasks. Commonly, features with high absolute gradient matters much in the prediction. As a result, people can use the gradients backpropagate through the model as information to have some intuitions about the behavior of the model. Methods based on gradients includes Guided Backprop\cite{GB}, CAM\cite{CAM}, Grad-CAM\cite{Grad-CAM}, Grad-CAM++\cite{Grad-CAM++} and so on. In Guided Backprop, people make use of the gradients backpropagated through the model, and then apply gradient ascent algorithm to a certain layer or neuron to maximize the value of it. After that people use these values to visualize the layer to get some insights regarding what this layer is patterning. In CAM, GradCAM, GradCAM++, people use GAP\cite{CAM}(global average pooling) in the last fully conneted layer. And use the learned weights as the contribution corresponding to each class activation map. And finally, having get the synthesized image related to the weights and class activation map, people can in some sense understand why the model predicts so.

\section{Conclusion and future work}

In this paper, we propose a new model-agnostic explanation method called CLE, an approach to explain the predictions of any classifier in an interpretable manner. In the process of perturbation, CLE takes contextual information into account, which leads to an explanation which contains a variety of meaningful correlations. Our experiments demonstrated that CLE is more faithful to the black box model when compared to LIME. And CLE explanation is both reliable and helpful to users.

The computational load of CLE is relatively high, although users can pass in the features $E^\lambda$ they care about to make it lower. In the future, we will work to optimize the calculation process, to provide users an accurate and real-time explanation.

\end{document}